\documentclass[conference]{IEEEtran}
\IEEEoverridecommandlockouts

\usepackage[linesnumbered,ruled,vlined]{algorithm2e}
\usepackage{cite}
\usepackage{amsmath,amssymb,amsfonts}
\usepackage{algorithmic}
\usepackage{graphicx}
\usepackage{textcomp}
\usepackage{xcolor}
\usepackage{bm}
\usepackage{bbm}
\usepackage{color,soul}
\usepackage{adjustbox}
\usepackage{threeparttable}
\usepackage{graphicx}
\usepackage{booktabs}
\usepackage{multirow}
\graphicspath{ {./figs/} }

\def\BibTeX{{\rm B\kern-.05em{\sc i\kern-.025em b}\kern-.08em
    T\kern-.1667em\lower.7ex\hbox{E}\kern-.125emX}}

\newtheorem{definition}{Definition}
\newtheorem{assumption}{Assumption}
\newtheorem{lemma}{Lemma}
\newtheorem{theorem}{Theorem}

\begin{document}

\title{
Metric-Free Individual Fairness with Cooperative Contextual Bandits
}

\author{
\IEEEauthorblockN{Qian Hu, Huzefa Rangwala}
\IEEEauthorblockA{Department of Computer Science, George Mason University} 
\{qhu3, rangwala\}@gmu.edu

}

\maketitle

\begin{abstract}
Data mining algorithms are increasingly used in automated decision making 
across all walks of daily life. Unfortunately, as reported in several studies 
these algorithms learn bias from data and environment leading to unequitable and unfair 
solutions. 
%
%
To mitigate bias in machine learning, different 
formalizations of fairness have been proposed that 
can be categorized into group fairness and individual 
fairness. Group fairness requires that different groups should be 
treated similarly  which might be unfair to 
some individuals within a group. On the other hand, 
individual fairness requires that similar individuals be treated similarly.
%
However, individual fairness 
remains understudied due to its reliance on 
problem-specific similarity metric.
We propose a metric-free 
individual fairness and a
cooperative contextual bandits (\textbf{CCB}) algorithm. The \textbf{CCB} algorithm
utilizes  fairness as a reward and attempts to 
maximize it. The advantage of treating
fairness as a reward is that the fairness criterion does not need to
be differentiable.
The proposed algorithm is tested on multiple real-world benchmark
datasets. The 
results show the effectiveness of the proposed algorithm at 
mitigating bias and at achieving both 
individual and group fairness.
\end{abstract}


\section{Introduction}
Machine learning  (ML) algorithms have 
permeated many levels of human society and  
 achieved tremendous 
success in many scenarios ranging from improved healthcare, education 
and commerce. 
For example, ML enables people with disabilities to 
live better lives \cite{zhao2015foresee,rudovic2018personalized} and 
can identify tumors before the on-set of cancer 
\cite{mckinney2020international}. Educational technologies based on
ML  provide personlalized learning experiences to students and 
improve their learning outcomes \cite{baker2014educational}. 

However, concerns and evidence
of machine learning bias are increasing 
with its adoption. For example, 
Buolamwini et al. \cite{pmlr-v81-buolamwini18a} found that commercial face recognition
software exhibit skin-type and gender biases. This study showed that the classification error 
rate for light-skinned men is substantially lower than dark-skinned women, 0.8\% and 34.7\%, respectively. A recidivism prediction algorithm COMPAS investigated by ProPublica misclassifies African-American defendants at a higher risk of recidivism than their Caucasian counterparts (45\% vs. 23\%) \cite{propublica_mb}. A study by Ali et al. \cite{ali2019discrimination} showed skewed job and housing advertisement delivery. For instance, advertisements for jobs in taxi industry disproportionately target communities of color (75\%), even though the specified targeting audience is identical for all race. 
The increasing amount of bias cases in machine learning applications require the development of bias-free algorithms.

Therefore, research communities in 
machine learning, policy, social science, law and 
statistics have invested in the development of fairness definitions
\cite{article_Pratik_formalizing}, and methods \cite{JMLR:v20:18-262,NIPS2017_6995,Zafar_2017,calders2010three}. The proposed fairness definitions include fairness through unawareness \cite{dwork2012fairness}, counterfactual fairness \cite{NIPS2017_6995}, statistical parity \cite{feldman2014certifying}, equalized odds and equal opportunity \cite{DBLP:journals/corr/HardtPS16} and individual fairness \cite{dwork2012fairness}.

The proposed fairness definitions can be classified into two categories: group and individual fairness. Group fairness requires that the protected group as a whole should be treated statistically similarly to the advantaged groups. For example, if the hiring rate for 
males is 60\%,  then the hiring rate for female candidates should be 60\% as well. 
Although most existing algorithms proposed for achieving fairness focus on group fairness, it has the potential to be unfair to specific individuals. For example, we can accept certain
people in an identified 
group to achieve group fairness, as long as the ratio of 
being accepted is identical across all groups. However, it is unfair 
to qualified people in that group. In addition, forcing the
acceptance rates to be the same across all 
groups can lead to inferior prediction performance if the distribution across the different 
groups are different. 

As an alternate formalization,   individual fairness 
was proposed by Dwork et al. \cite{dwork2012fairness}. This 
requires that similar individuals be treated similarly. To implement this concept, a problem-specific similarity metric is required. However, defining a similarity metric for each task has its own ethical issues, which is a critical challenge in  
making individual fairness practical \cite{DBLP:journals/corr/abs-1810-08810}. 

We propose the concept of metric-free individual fairness where 
the prediction of an individual should not be influenced by their sensitive attribute. Compared to 
traditional individual fairness which assures fairness by comparing an individual to others, the 
proposed metric-free individual fairness compares the predictions of an individual conditioned 
on different values of the sensitive attribute.

To implement the proposed metric-free individual fairness, we propose an
algorithm called
cooperative contextual bandits. The algorithm consists of 
multiple contextual bandits where each bandit corresponds to a
sensitive attribute. Given an individual, the algorithm predicts
the individual's probability of getting
a positive outcome using the bandit corresponding to 
their sensitive attribute and compares it to the probability of 
getting a positive outcome by using the bandit corresponding to the other 
sensitive attribute value(s).
The proximity of the two probability distributions is treated as fairness. Namely, if the prediction of the 
individual is not affected by their
sensitive attribute, it is considered fair. The algorithm treats the proximity of the 
two probability distributions as a reward signal and 
maximizes it to achieve fairness. As we need to measure the divergence of the predictions between different bandits, we develop a gradient contextual bandits algorithm. Compared to traditional contextual bandits which predict the action value, gradient contextual bandits directly map a state to an
action distribution that can be used to measure divergence.

The proposed algorithm is analyzed theoretically and empirically. It is shown that the proposed fair algorithm converges linearly to a local minimum. We empirically evaluate the algorithm on several public datasets and the experimental results show the effectiveness of the proposed algorithm at mitigating bias. 

The contributions of this work can be summarized as follows
\begin{itemize}
	\item We propose a new concept of individual fairness which does not need a similarity metric. The proposed metric-free individual fairness allows for broader adoption of individual fairness.
	\item We propose gradient contextual bandits algorithm. 
	The proposed gradient contextual bandits can be used in many other domains such as recommender systems.
	One of the advantage of the gradient contextual bandits is that it learns a stochastic policy, while the policy learned by classic contextual bandits is deterministic. 
	\item Based on gradient contextual bandits, we propose a novel algorithm, cooperative contextual bandits, to implement the proposed metric-free individual fairness. The proposed algorithm relies on the reward signal to achieve fairness, therefore, the fairness measure is not required to be differentiable.
\end{itemize}


\section{Related Work} \label{sec:related_work}

\subsection{Fairness Formalization}
Different definitions of fairness have been proposed \cite{mehrabi2019survey}. 
\textbf{Demographic parity} imposes the constraint that a predictor predicts a particular outcome for  individuals across groups with nearly equal probability \cite{feldman2015certifying,corbett2017algorithmic,barocas2016big}. This notion has the disadvantage of trading utility for fairness. 
Hardt et al. \cite {hardt2016equality} proposed \textbf{equalized odds} and \textbf{equal opportunity}. A classifier satisfies equalized odds if the predicted outcome is independent of the sensitive attribute conditioned on the true outcome. For binary classification, if the true outcome is positive, equalized odds requires that a classifier yield 
equal true positive rates across the groups. If the true outcome is negative, it requires the predictions lead to 
equal false positive rates. Equalized odds is a stronger
constraint compared to equal opportunity. Equal opportunity only requires non-discrimination within the advantaged group. For example, the qualified students in different groups should have similar probability of getting admitted into colleges.

Based on causal inference, Kusner et al. \cite{NIPS2017_6995} proposed \textbf{counterfactual fairness}. A predictor satisfies counterfactual fairness if its output remains the same when an individual's sensitive attribute is flipped to its counterfactual value. Counterfactual fairness is an individual-level fairness formalization.
Dwork el al. \cite{dwork2012fairness} explicitly proposed \textbf{individual fairness} based on the idea that similar individuals should be treated similarly. However, this notion of fairness largely depends on the similarity metric. To achieve this notion of fairness, a reliable and non-discriminating similarity metric is required \cite{article_Pratik_formalizing}.

Recently, several efforts have been made to eliminate the need of similarity metric of individual fairness \cite{gitiaux2019multi}. Ilvento \cite{ilvento2019metric} proposed to approximate a similarity metric for individual fairness by using human judgement, namely, a human provides with sufficient domain knowledge to evaluate similarity. Although an explicit similarity metric is not necessary in this work, human judgements are needed, which is prone to error and hard to implement.
Bechavod et al. \cite{bechavod2020metric} proposed a metric-free online learning algorithm which assumes access to a human
auditor identifying fairness violations. 


Different from the aforementioned metric-free individual fairness, our definition of metric-free individual fairness needs no human judgements.

\subsection{Fair Algorithms}
In this section, we mainly survey the works on individual fairness.
Based on John Rawls' notion of fair equality of opportunity, 
Joseph et al. \cite{joseph2016fairness} proposed Rawlsian fairness, which states that in the case of hiring, a weaker 
applicant should never be favored over a 
stronger candidate. Their proposed fair algorithm is based on contextual bandits and confidence intervals, which chains 
confidence intervals of different individuals to determine which individuals should be favored over others.
Zemel et al. \cite{zemel2013learning} take a different perspective on how to achieve individual fairness. They proposed an algorithm to learn a representation while removing sensitive information from the learned representation. The proposed algorithm is able to achieve both individual and group fairness.

Following the idea of learning a fair representation, Edwards et al. \cite{edwards2015censoring} proposed to remove sensitive information from the representation by adversarial learning. An encoder learns to generate the representation and an adversary predicts the sensitive attributes from the representation. To achieve fairness, the adversary tries to predict sensitive information from the representation while the encoder seeks to refrain the adversary from predicting it.
Louizos et al. \cite{louizos2015variational} proposed using variational autoencoder for learning a fair representation by encouraging independence between sensitive attributes and the representation. To further remove dependencies, the algorithm is integrated with an additional term based on maximum mean discrepancy.
Compared to the algorithms achieving individual fairness by learning a fair representation, our algorithm directly optimizes the individual fairness criterion.

\subsection{Cooperative Contextual Bandits}
Cooperative contextual bandits have been applied in areas ranging from recommender systems \cite{cesa2013gang,wu2016contextual,kolla2018collaborative} to robotics \cite{Landgren2019Distributed}. Although it is based on the idea of combining multiple bandits, the ways that how the bandits are combined and the reasons to combine them are different.

To improve information sharing and overcome data sparsity issues in recommender systems where users are connected through social networks, Cesa et al. \cite {cesa2013gang} proposed to combine the bandits through the Laplacian matrix of the graph. A similar work done by Wu et al. \cite {wu2016contextual} combines the bandits by parameters propagation through the graph. In robotics, Landgren \cite {Landgren2019Distributed} proposed distributed multi-agent bandits for searching robots which are combined by using consensus communication protocol. Each robot is represented by a bandit, and the main reason for combining multiple bandits is to help agents with sensor defects.

Our proposed cooperative contextual bandits algorithm is different from existing algorithms in the sense that we impose constraints on the output space of the bandits.

\section{Preliminaries} \label{sec:preliminaries}
In this section, we present traditional definition of individual fairness and contextual bandits. 
Without loss of generality, in this work we assume a binary sensitive attribute.

Given an individual $((x_i, s_i), y_i)$, where $x_i \in X$ denotes the feature vector of individual $i$, $s_i \in \{ 0, 1 \}$ is the individual's sensitive attribute, $y_i$ is the ground-truth label.

\subsection{Metric-Based Individual Fairness}
The most popular individual fairness is based on the idea that similar individuals should be treated similarly, regardless of their sensitive attributes. Given a classifier, two individuals $x_i$ and $x_j$ are mapped to two distributions $H(x_i)$ and $H(x_j)$, respectively. Individual fairness requires that the difference between the 
two output distributions is upper bounded by the distance between the two individuals.
\begin{definition}
(Metric-Based Individual Fairness)
$\forall x_i, x_j \in X $, the metric based individual fairness has the constraint that $ D(H(x_i), H(x_j)) \leq d(x_i, x_j)$, where $d: X \times X \to \mathbb{R} $ is a distance metric for two individuals and $D$ measures divergence between the two outcomes. 
\end{definition}


\subsection{Contextual Bandits}
The classic contextual bandits problem is formalized as a finite set of $K$ actions $A$, context space $X$, a set of policies $\Pi$ which maps a context $x \in X$ to an action $a \in A$ based on action value $Q(x, a)$.

At time $t = 1,...,T$, a context $x_t$ is shown. A policy chooses an action $a_t$ by estimating reward value of actions based on the history 
$(x_1, a_1, r_1(a_1)), ..., (x_{t-1}, a_{t-1}, r_{t-1}(a_{t-1}))$ and receives a reward $r_t(a_t)$. The goal of the contextual bandit algorithm is to maximize the expected reward of a policy $R(\pi) = E_{(x, r) \sim D}[r(\pi(x))] = \sum_{t=1}^{T}r_t(a_t)$, where $D$ is the distribution over context and reward.

At each round, the policy decides which action to take. It has two options, taking the action with the highest estimated value (exploitation) or explore actions that might have higher value (exploration). Contextual bandit algorithms have to make a sensible tradeoff between exploitation and exploration. Many algorithms have been proposed to overcome the exploitation and exploration challenge such as $\epsilon$-greedy, LinUCB \cite{li2010contextual} and Thompson Sampling \cite{agrawal2013thompson}.

\section{Proposed Algorithm} \label{sec:main_algo}
In this section, we propose our algorithms for metric free individual fairness based on gradient contextual bandit.
\subsection{Metric-Free Individual Fairness} \label{sec_def_metric_free}

\begin{definition}
(Metric-Free Individual Fairness)
Given an individual with $(x_i, s_i)$, metric-free individual fairness is defined as, $D(P(Y|x_i, S = s_i), P(Y|x_i, S = 1 - s_i)) < \epsilon $, for arbitrarily small positive constant $\epsilon$, where $P(Y|x_i, S = s_i)$ is the distribution of the outcome given an individual $x_i$ and the sensitive attribute $s_i$.
\end{definition}

Metric-free individual fairness requires that the outcome of an individual is
independent of their
sensitive attributes given their input features. This  
implies that the prediction for an individual should
not be influenced by their sensitive attributes.

\subsubsection{Relationship to Counterfactual Fairness}
Counterfactual fairness \cite{NIPS2017_6995} is defined such
that a predictor $\hat{Y}$ is counterfactually fair for any individual $x$ with sensitive attribute $s$, if $P( \hat{Y}_{S \leftarrow s} | X = x, A = a) = P( \hat{Y}_{S \leftarrow s'} | X = x, A = a)$, where $S \leftarrow s$ denotes the intervention that sets sensitive attribute $S$ to value $s$.

The main difficulty of adopting counterfactual fairness is untestable  
assumptions about causal relationships between 
variables \cite{russell2017worlds}. Although counterfactual fairness holds under different levels of assumptions 
from weak to strong, the utility and feasibility of the models under these assumptions decrease \cite{NIPS2017_6995}. For example, the 
weakest assumption only requires to identify non-descendants of sensitive attribute, but in many practical problems there exist few non-descendants of sensitive attribute. The strongest assumption postulates a fully deterministic model which guarantees the
maximum information extraction, however, it is the hardest to validate. Validating a causal assumption is sometimes unrealistic; Russell et al. \cite{russell2017worlds} propose to avoid doing this by integrating multiply competing assumptions. However, it results in approximate counterfactual fairness and still requires the formalization of causal relations between variables.
Compared to counterfactual fairness, the proposed metric-free individual fairness is based on conditional independence between the prediction and the sensitive attribute, which requires no assumptions and is simple to implement. The experimental results show the effectiveness of models implementing it at removing bias.

\subsection{Gradient Contextual Bandit}
Contextual bandits algorithms make a decision at each step by estimating the
action value $Q(x, a)$, which is the expected reward of taking action $a$ given context $x$. The 
gradient contextual bandits 
learns a stochastic policy that maps a context vector to a probability distribution of actions. This is desirable in our case as we want to know the distance between two distributions of actions given a context vector, which is detailed in Section \ref{sec_reward_function}. Besides that, a stochastic policy is beneficial in other domains such as recommender systems. For example, a user might like an action movie as much as a science fiction movie, which can be easily modeled by using a stochastic policy, while a deterministic policy can only choose the movie that the user likes the best.

Formally, we want to learn a policy $\pi_{\theta}(a | x) = P(A = a | X = x, \theta)$ parameterized by $\theta$, which is the probability of choosing action $a$ given context vector $x$ that maximizes the expected reward
\begin{equation}
J(\theta) = E_{x \sim P(X)}[\pi_{\theta}(a | x)Q(x, a)]
\end{equation}
In the following text, we ignore $\theta$ and use $\pi_{\theta}(a | x)$ and $\pi(a | x)$ interchangeably. As the action space is discrete, we can parameterize the policy as
\begin{equation} \label{eq:policy_para}
	\pi(a|x) = \frac{e^{\phi(x,a)}}{\sum_b e^{\phi(x,b)}}
\end{equation}
where $\phi(x, a)$ is a preference score of choosing action $a$, which is mapped from feature vector $x$.

The gradient of the expected reward of a policy with respect to the policy parameter is
\begin{equation}
	\nabla J(\theta) = E_{x \sim P(X)}[\nabla \pi_{\theta}(a | x)Q(x, a)]
\end{equation}
However, the probability distribution of X is unknown. Instead, we seek to optimize the empirical reward $J_t(\theta)$, which is an unbiased estimator of the expected reward, namely, $J(\theta) = E[J_t(\theta)]$.
The empirical policy gradient is 
\begin{equation}
	\nabla J_t(\theta) = \nabla \log ( \pi_{\theta} (a_t|x_t)r_t(a_t) ).
\end{equation}
The gradient of the policy with respect to its parameters is the gradient of the logarithm of the policy times the reward. The derivation of the policy gradient is detailed as following
\begin{equation}
	\begin{aligned}
	\nabla J_t(\theta) & = \sum_a \nabla \pi_{\theta}(a | x_t)q(x_t, a) \\
	& = \sum_a \pi_{\theta}(a | x_t) \frac{\nabla \pi_{\theta} (a | x_t)}{\pi_{\theta} (a | x_t)} q(x_t, a) \\
	& = E_{\pi_{\theta}}[ \nabla \log \pi_{\theta} (a | x_t) q(x_t, a) ] \\
	& = E_{\pi_{\theta}}[ \nabla \log \pi_{\theta} (a_t | x_t) q(x_t, a_t)] \\
	& = E_{\pi_{\theta}}[ \nabla \log \pi_{\theta} (a_t | x_t)R_t ] \\
	& = \nabla \log \pi_{\theta} (a_t | x_t)R_t. \\
	\end{aligned}
\end{equation}
where $E_{\pi_{\theta}}$ is the expectation taken with respect to $\pi_{\theta}$. 
Interestingly, we find that the derived algorithm is similar to the REINFORCE algorithm \cite{williams1992simple}. The difference between contextual bandits and full reinforcement learning (RL) is that the action taken by full RL agent will influence the environment while that taken by a contextual-bandit agent will not. Therefore, REINFORCE uses return from current step, while gradient contextual bandits use reward at current step.

The policy parameter can be updated by using stochastic gradient ascent algorithm as
\begin{equation}
	\theta_{t+1} = \theta_t + \alpha \nabla \log \pi_{\theta}(a_t|x_t)r_t(a_t).
\end{equation}

\subsection{Cooperative Contextual Bandits}

\begin{figure} [ht!]
	\centering
	\includegraphics[width=0.95\linewidth]{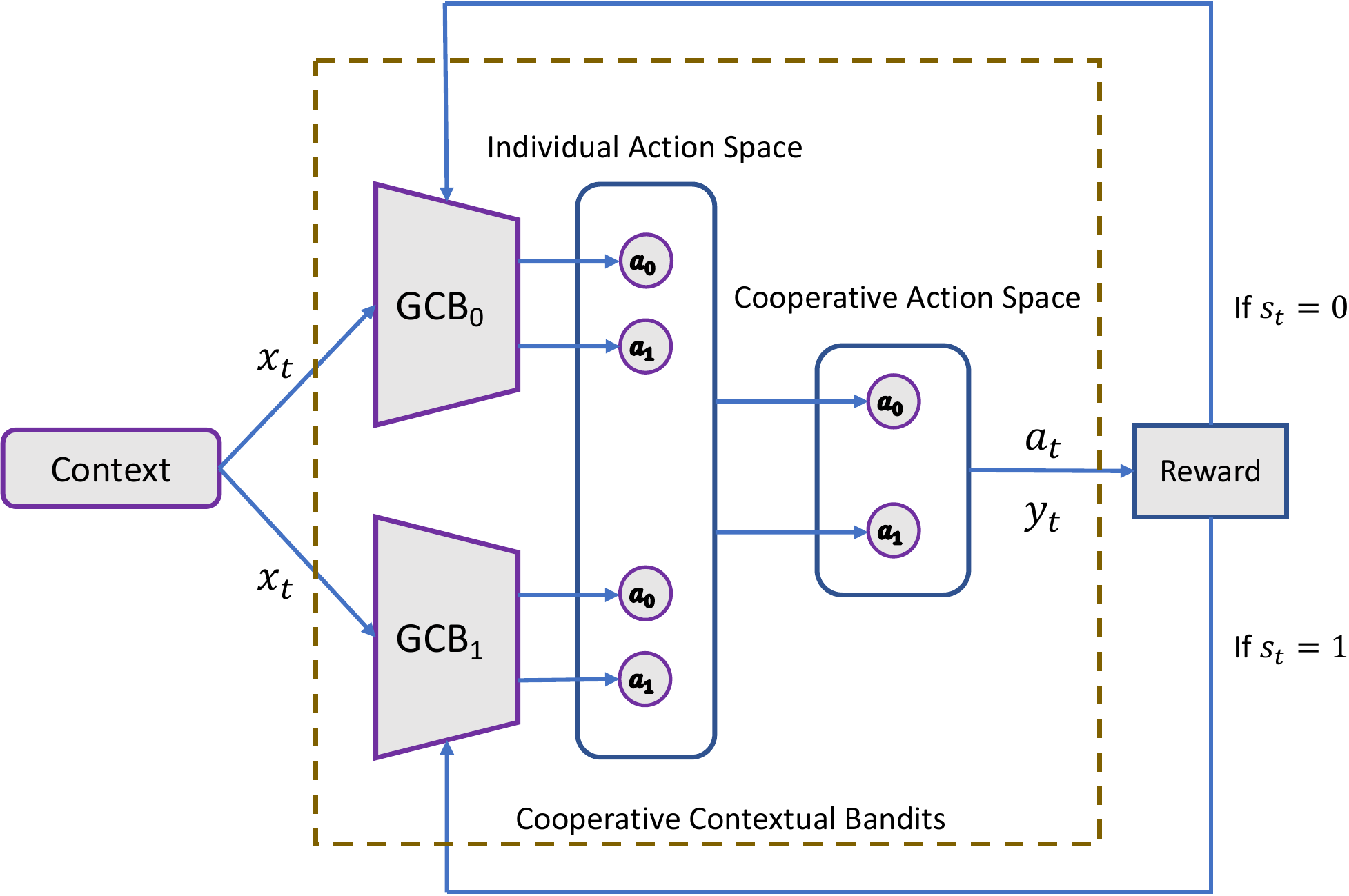}
	\caption{The Cooperative Contextual Bandits Model. GCB\textsubscript{0} and GCB\textsubscript{1} represent the bandits corresponding to sensitive attribute values of 0 (male) and 1 (female), respectively.} \label{fig:model_arch}
\end{figure}


Metric-free individual fairness imposes the constraint that the prediction for an individual is independent of their sensitive attribute. We propose cooperative contextual bandits algorithm which is composed of two gradient contextual bandits. Figure \ref{fig:model_arch} shows the model architecture of the proposed model.
Each gradient contextual bandit corresponds to a sensitive attribute group, e.g., male or female. Each gradient contextual bandit learns a policy $\pi_s$ which maps a feature vector to an action distribution.
The individual's context vector is  denoted by $x_t$ and 
is input into both bandits, each of which outputs an action distribution. The algorithm takes an action according to the action distribution of the bandit corresponding to $s_t$. A reward is determined by the ground truth label $y_t$ and the divergence between the action distributions of the two bandits. The bandit corresponding to $s_t$ receives the reward feedback and gets updated. Note that the other bandit is not updated.


According to the definition of metric-free individual fairness, the difference between outcomes given by different gradient contextual bandits should be trivial, i.e. $D(\pi_{s_i}(x_i), \pi_{1 - s_i}(x_i)) < \epsilon$, for arbitrarily small $\epsilon$. Since we are using gradient contextual bandits which directly maps the context vector to the probability distribution of actions, the difference of the outcomes can be measured by KL divergence between the outcomes. Note that, metric-free individual fairness in this work means that there is no requirement of similarity metric between individuals, while we still need to measure the divergence between the two output distributions. The former is problem-specific and hard to justify if it is a fair metric, while the latter is easy to define and the choice of which has no effect on fairness.

\subsubsection{Reward Function} \label{sec_reward_function}
At time step $t$, an individual $(x_t, s_t)$ is presented, an action $a_t$ is taken and a reward $r_t(a_t)$ is received. The action is the predicted outcome for the individual, i.e., whether he/she is hired. The reward function is composed of two parts, fairness and accuracy.
\begin{equation}
	r_t(a_t) = r_t^a(a_t) - \lambda \text{KL} (\pi_{s_t}(x_t), \pi_{1 - s_t}(x_t)) \label{reward_func}
\end{equation}
where $r_t^a(a_t) \in \{ 0, 1 \}$ is the accuracy reward,
\[
	r_t^a(a_t) = 
\begin{cases}
	0, & a_t \neq y_t \\
	1, & a_t = y_t
\end{cases}
\]
$\lambda$ is a hyperparameter trading off between accuracy and fairness. The reward is high when the prediction is correct while the difference between the outcomes from different contextual bandits is small.

\subsubsection{Model Architecture of Gradient Contextual Bandits}
The input into the gradient contextual bandit is a feature vector describing an individual. The bandit learns a policy that maps the feature vector to an action distribution.
The policy is parameterized as a feed-forward neural network. As we assume a binary discrete label, the last layer of the neural networks is a softmax layer whose output is the probability distribution of actions. The neural networks are composed of one hidden layer and the activation function is a ReLU function \cite{hahnloser2000digital}.

Algorithm \ref{main_alg} shows the detail of the proposed algorithm \footnote{The code is made public at https://github.com/anony-user-123/public}.
\begin{algorithm} 
	\caption{Cooperative Contextual Bandits} \label{main_alg}
	\DontPrintSemicolon
	\SetKwInOut{Input}{Input} \SetKwInOut{Output}{Output}
	\SetKw{Initialize}{Initialize}

	\Input{Data $D = \{ ((x_i, s_i), y_i) \}_{i=1}^{N}$, learning rate $\alpha$, $\lambda$, number of steps T.}
	\BlankLine

	Initialize parameters $\{ \theta_0^0, \theta_1^0 \}$

	\For{$t = 1, ..., T$} {
		Sample example $((x_t, s_t), y_t)$ from $D$ \;
		Sample $a_t \sim \pi_{s_t}(x_t)$ \;
		Take action $a_t$, compute reward $r_t(a_t)$ using Equation \ref{reward_func} \;
		$ \theta_{s_t}^{t+1} = \theta_{s_t}^{t} + \alpha \nabla \log B_{s_t}(a_t | x_t) r_t(a_t) $
	}
	\KwRet{$\{ \theta_0^{T}, \theta_1^{T} \}$}
\end{algorithm}

\subsection{Convergence Analysis}
In this section, we analyze the convergence property of the proposed algorithm.

\begin{assumption} \label{asu_reward}
There exists $Q$, such that the reward function satisfies
\begin{equation}
	|r_t(a_t)| \leq Q
\end{equation}
for all $t$ and $a_t$.
\end{assumption}
Assumption \ref{asu_reward} states that the reward function is bounded.

\begin{assumption} \label{assumption2}
Let $s \in \{ 0, 1 \}$. The first and second derivatives of the score function are elementwise bounded by constants $F$ and $S$, respectively, and $0 \leq F, S < \infty$
\begin{equation}
	| \nabla \log \pi_{s} | \leq F
\end{equation}
\begin{equation}
	| \nabla^2 \log \pi_{s} | \leq S.
\end{equation}
\end{assumption}

The policy is parameterized as
\begin{equation}
\pi_{s}(a | x) = \frac{e^{\phi(x, a)}}{\sum_b e^{\phi(x, b)}}.
\end{equation}
Therefore, $\nabla \log \pi_{s} = (1 - \pi_s(a|x)) \nabla \phi(x, a)$, which is bounded if $\nabla \phi(x, a)$ and the parameter $\theta_s$ is bounded.  The second derivative is derived as
\begin{equation}
\begin{aligned} 
	\nabla^2 \log \pi_s(a|x) & = (1 - \pi(a|x))^2 \nabla^2 \phi(x, a) \nabla^2 \pi_s(a | x) \\
	& - (1 - \pi(a|x))^2 \nabla \phi(x,a) \nabla \phi(x,a)^T
\end{aligned}
\end{equation}
which is also bounded if $\nabla^2 \phi(x, a)$, $ \nabla^2 \pi_s(a | x)$ and $\nabla \phi(x,a)$, $\theta_s$ are bounded.

\begin{lemma} \label{lemma1}
Under Assumption \ref{assumption2}, for $ \forall s \in \{ 0, 1 \}, J_{s}(\theta)$ is $L$-Lipschitz smooth and $0 \leq L < \infty$
\begin{equation}
|| \nabla J_{s}(\theta_1) - \nabla J_{s}(\theta_2) || \leq L || \theta_1 - \theta_2 ||
\end{equation}
\end{lemma}
Lemma \ref{lemma1} assures that the objective function $J_s(\theta)$ is smooth, which is essential in analyzing the convergence of the algorithm. 
\paragraph{Proof} 
The Hessian matrix of the score function $\log \pi_{\theta_{s}}$ is
\begin{equation}
\nabla^2 \log \pi_{\theta_s} = \pi_{\theta_s}^{-1} \nabla^2 \pi_{\theta_s} - \nabla \log \pi_{\theta_s} \nabla \log \pi_{\theta_s}^T.
\end{equation}
Therefore,
\begin{equation}
\nabla^2 \pi_{\theta_s} = \pi_{\theta_s} \nabla^2 \log \pi_{\theta_s} + \pi_{\theta_s} \nabla \log \pi_{\theta_s} \nabla \log \pi_{\theta_s}^T.
\end{equation}
The second derivative of the objective function is bounded as shown below
\begin{equation}
\begin{aligned}
\nabla^2 J(\theta_s) & = \int_{X} \nabla^2 \pi_{\theta_s}(a | x)q(x, a) d x \\
& = \pi_{\theta_s} ( \int_{X} \nabla^2 \log \pi_{\theta_s}(a|x)q(x, a) d x + \\
& \int_{X} \nabla \log \pi_{\theta_s}(a|x) \nabla \log \pi_{\theta_s} (x | a)^T q(x, a) d x ) \\
& \leq SQ + F^2 Q = L. \\
\end{aligned}
\end{equation}
Bounded second derivative implies Lipschitz continuity, therefore $J(\theta_s)$ is $L$-smooth.

\begin{theorem} \label{theorem}
For $ \forall s \in \{ 0, 1 \}$, let $\{ \theta_s^t \}_{t=0}^{T}$ be the sequence of learned parameters of the policy $\pi_{s}$ given by Algorithm \ref{main_alg}. Then, under Assumption \ref{asu_reward} and \ref{assumption2}, we have
\begin{equation}
|| \nabla J(\theta_{s}^m) ||_2^2 \leq \frac{1}{T} \sum_{t=0}^{T} || \nabla J(\theta_{s}^k) ||_2^2 \leq \frac{J(\theta_{s}^*) - J(\theta_{s}^0)}{MT},
\end{equation}
where $\nabla J(\theta_{s}^m) =  \underset{t \in \{ 0, ..., T \}}{\min} \nabla J(\theta_{s}^t)$, $J(\theta_{s}^*) =  \underset{t \in \{ 0, ..., T \}}{\max} J(\theta_{s}^t)$, and M is a constant with $0 < M < \infty$.
\end{theorem}

Theorem \ref{theorem} shows that the average of the gradient norm square converges to a neighborhood near zero with rate of $\frac{1}{T}$ or the minimizer of the gradient norm square converges to zero in $O (\frac{1}{T})$ iterations. The minimizer can be obtained by using early stopping. 
\paragraph{Proof}
The objective function $J(\theta_s)$ is smooth, which directly implies
\begin{equation}
\begin{aligned}
J(\theta_{s}^{k+1}) & \geq J(\theta_{s}^k) + \langle \nabla J(\theta_{s}^k), \theta_{s}^{k+1} - \theta_{s}^k \rangle  - \frac{L}{2} || \theta_s^{k+1} - \theta_s^k ||_2^2 \\
& \geq J(\theta_{s}^{k}) + \alpha || \nabla J(\theta_s^k) ||_2^2 - \frac{L}{2} || \theta_{s}^{k+1} - \theta_{s}^k ||_2^2 \\
& \geq J(\theta_s^k) + \alpha || \nabla J(\theta_s^k) ||_2^2 - \frac{L \alpha^2}{2} || \nabla J(\theta_s^k) ||_2^2 \\
& \geq J(\theta_s^k) + \frac{2\alpha - L\alpha^2}{2} || \nabla J(\theta_s^k) ||_2^2 \\
& \geq J(\theta_s^k) + M || \nabla J(\theta_s^k) ||_2^2 \\
\end{aligned}
\end{equation}
where $M = \frac{2\alpha - L\alpha^2}{2}$.

By telescoping sum the above equation with $k$ from $0$ to $K$, we obtain 
\begin{equation}
J(\theta_s^k) \geq J(\theta_s^1) + M \sum_{k=0}^{K} || \nabla J(\theta_s^k) ||_2^2.
\end{equation}

Therefore,
\begin{equation}
\begin{aligned}
\frac{1}{K} \sum_{k=0}^{K} || \nabla J(\theta_s^k) ||_2^2 & \leq \frac{J(\theta_s^k) - J(\theta_s^1)}{MK} \\
& \leq \frac{J(\theta_s^*) - J(\theta_s^1)}{MK}
\end{aligned}
\end{equation}
where $J(\theta_s^*)$ is the maximizer of $J(\theta_s^k)$ with respect to k.

Defining $\nabla J(\theta_{s}^m) =  \underset{t \in \{ 0, ..., T \}}{\min} \nabla J(\theta_{s}^t)$, we obtain
\begin{equation}
|| \nabla J(\theta_{s}^m) ||_2^2 \leq \frac{1}{T} \sum_{t=0}^{T} || \nabla J(\theta_{s}^k) ||_2^2 \leq \frac{J(\theta_{s}^*) - J(\theta_{s}^0)}{MT}.
\end{equation}
This Completes the proof for Theorem \ref{theorem}






\begin{table*}[h!]
	\centering
	\caption{Experimental results on public datasets using discrimination as model selection criterion.} \label{tab:rslt1}
	\begin{adjustbox}{max width=\textwidth}
	\begin{threeparttable}

\begin{tabular}{ccccc}
\toprule
{} &                 adult &                compas &                german &                health \\
{} & acc$|$discri$|$consist & acc$|$discri$|$consist & acc$|$discri$|$consist & acc$|$discri$|$consist \\
\midrule
\textbf{LR}       &  0.8387$|$0.1873$|$0.8307 &  \textbf{0.6878}$|$0.2859$|$0.7493 &  \textbf{0.7700}$|$0.0759$|$0.6590 &  \textbf{0.8786}$|$0.0119$|$0.9883 \\
\textbf{Rawlsian} &  0.8378$|$0.1686$|$0.8359 &  0.6802$|$0.3877$|$0.7054 &  0.7500$|$0.0287$|$0.6920 &  0.8765$|$0.0072$|$0.9837 \\
\textbf{LFR}      &  0.7548$|$0.0016$|$0.9990 &  0.6254$|$0.0356$|$0.5114 &  0.6950$|$0.0282$|$0.9840 &  0.8785$|$\textbf{0.0000}$|$\textbf{1.0000} \\
\textbf{ALFR}     &  \textbf{0.8420}$|$0.0722$|$0.8473 &  0.4986$|$0.0487$|$0.9101 &  0.6850$|$\textbf{0.0000}$|$\textbf{1.0000} &  0.8785$|$\textbf{0.0000}$|$\textbf{1.0000} \\
\textbf{CCB}      &  0.7544$|$\textbf{0.0002}$|$\textbf{0.9999} &  0.4768$|$\textbf{0.0031}$|$\textbf{0.9960} &  0.6950$|$0.0282$|$0.9840 &  0.8785$|$0.0002$|$0.9999 \\
\bottomrule
\end{tabular}

	\begin{tablenotes}
	\small
	\item acc = accuracy, discri = discrimination, consist = consistency.
	\end{tablenotes}
	\end{threeparttable}
	\end{adjustbox}
\end{table*}

\begin{table*}[h!]
	\centering
	\caption{Experimental results on public datasets using delta as model selection criterion.} \label{tab:rslt2}
	\begin{adjustbox}{max width=\textwidth}
	\begin{threeparttable}

\begin{tabular}{ccccc}
\toprule
{} &                 adult &                compas &                german &                health \\
{} & acc$|$discri$|$consist & acc$|$discri$|$consist & acc$|$discri$|$consist & acc$|$discri$|$consist \\
\midrule
\textbf{LR}       &  \textbf{0.8387}$|$0.1873$|$0.8307 &  \textbf{0.6878}$|$0.2859$|$0.7493 &  \textbf{0.7700}$|$0.0759$|$0.6590 &  0.8786$|$0.0119$|$0.9883 \\
\textbf{Rawlsian} &  0.8373$|$0.1651$|$\textbf{0.8351} &  0.6802$|$0.3877$|$0.7054 &  0.7500$|$\textbf{0.0287}$|$0.6920 &  0.8765$|$0.0072$|$0.9837 \\
\textbf{LFR}      &  0.7756$|$0.0197$|$0.7938 &  0.6556$|$0.0408$|$0.5147 &  0.7500$|$0.1058$|$0.7020 &  0.8785$|$\textbf{0.0000}$|$\textbf{1.0000} \\
\textbf{ALFR}     &  0.8327$|$0.0465$|$0.8220 &  0.5137$|$0.0398$|$\textbf{0.9396} &  0.7450$|$0.0797$|$0.7760 &  \textbf{0.8803}$|$0.0015$|$0.9879 \\
\textbf{CCB}      &  0.8165$|$\textbf{0.0077}$|$0.8341 &  0.6254$|$\textbf{0.0145}$|$0.7272 &  0.7000$|$0.0423$|$\textbf{0.9810} &  0.8790$|$0.0021$|$0.9956 \\
\bottomrule
\end{tabular}

	\begin{tablenotes}
	\small
	\item acc = accuracy, discri = discrimination, consist = consistency.
	\end{tablenotes}
	\end{threeparttable}
	\end{adjustbox}
\end{table*}

\section{Experimental Protocol} \label{sec:experiment_protocol}
\subsection{Datasets}
We evaluate the performance of \textbf{CCB} model on four widely used benchmarks. We also perform a case study on  
a dataset collected at an educational institute for assessing 
the performance of various classifiers in academic performance prediction. All the tasks on these datasets are binary classification problem and the sensitive attributes are converted into two groups, i.e. protected and advantaged group. All the datasets are randomly split into 70\%, 15\% and 15\% for training, validation and testing, respectively. The categorical features are one-hot encoded and continuous features are normalized with z-score.
\paragraph{\textbf{Adult}} The Adult income dataset has 48,842 samples \cite{Dua:2019}. The goal is to predict if an adult's income is above 50K based on 14 categorical and continuous features. An individual's gender is the sensitive attribute.
\paragraph{\textbf{Compas}} The COMPAS algorithm is used to assess a pretrial defendant's risk of recidivism in Broward County, Florida. ProPublica collected 11,757 defendants assessed by COMPAS algorithm over two year from 2013 to 2014. Profiles were created for defendants using their criminal history, before and after they were scored by COMPAS \cite{propublica_mb}. The task is to predict whether a defendant will re-offend and the sensitive attribute is whether a defendant is black.
\paragraph{\textbf{German}} The German dataset has 1000 samples and 20 features \cite{Dua:2019}. The task is to predict whether a individual has a good or bad credit rating. The sensitive attribute is a person's gender.
\paragraph{\textbf{Health}} The Health dataset is extracted from Heritage Health Prize which has 147,743 samples. We follow Brierley et al. \cite{brierley2011heritage} to preprocess the dataset. The task of this dataset is to predict whether an individual will spend any days in the hospital in the next year. The sensitive attribute is a person's age. The age is converted to a binary variable, i.e., whether a persons is older than 65 years old.

\subsection{Baselines}
We choose several algorithms as baselines which focus on achieving individual fairness. 
\paragraph{Logistic Regression \textbf{LR}} This baseline does not have fairness constraint. \
\paragraph{Rawlsian Fairness \textbf{Rawlsian}} Joseph et al. \cite{joseph2016fairness} proposed a concept of individual fairness based on the idea that a worse candidate should never be favored over a better one. The algorithm is implemented by using classical contextual bandits. Given a pool of candidates, their context vectors are mapped to corresponding qualification intervals. Overlapping intervals are chained together and the candidates chained to the candidate with the highest upper confidence bound are treated equally and over other candidates. The hyperparameters for this model are regularization parameter $\lambda$ selected from $\{ 1.0, 3.0, 9.0 \}$ and the scaling parameter $\delta$ is chosen from $\{ 0.3, 0.5, 0.9 \}$.
\paragraph{Learning Fair Representation \textbf{LFR}} Zemel et al. \cite{zemel2013learning} proposed to train a fair representation and build a classifier based on the representation. The idea of making the representation fair is to remove sensitive information from it, which is achieved by imposing the constraint that a random element from the protected group has the same probability of mapping to a particular prototype as a random element from the advantaged group. Following the original work, we set $A\_x$ as 0.01, $A\_y$ and $A\_z$ are chosen from $\{ 0.1, 0.5, 1, 5, 10 \}$, the dimension of the representation is chose from $\{ 10, 20, 30 \}$.
\paragraph{Adversarial Learned Fair Representation \textbf{ALFR}} Following the idea of fair representation learning, Edwards et al. \cite{edwards2015censoring} proposed to learn a fair representation by using an adversary. The feature space is transformed to latent space by an encoder. An adversary tries to predict the sensitive attribute from the representation, while the encoder tries to generate a representation based on which the sensitive attribute can not be easily predicted. The encoder and the adversary play a minimax game to generate a fair representation. The hyperparameters of this method include the dimension of the representation which is chosen from $\{ 10, 20, 30, 40 \}$; and the number of neurons of the hidden layer which is chosen from $\{ 10, 20, 30, 40 \}$.

For the \textbf{CCB} algorithm, the hyperparameters include the number of neurons of the hidden layer and $\lambda$.
The number of neurons of the hidden layer is chosen from $\{ 10, 20, 30, 40 \}$ and $\lambda$ is chosen from $\{ 10, 20, 30, 40, 50, 60, 70, 80, 90, 100 \}$. 

\subsection{Evaluation Metrics} \label{sec:evaluation_metrics}
Following Zemel \cite{zemel2013learning}, we use three evaluation metrics \textbf{accuracy}, \textbf{discrimination} and \textbf{consistency} to evaluate the performance of the models.

The accuracy metric assesses the predictive accuracy of a classifier. Discrimination measures the difference between the two groups' rate of being predicted as positive. Consistency compares an individual's predicted result with his/her $k$-nearest neighbors. If the predicted result is close to the results of the neighbors, it has high consistency. We compute the similarity between individuals by using Gower's similarity \cite{gower1971general}. Gower's similarity can handle feature vectors with both categorical and continuous features.
Gower similarity is defined as
\begin{equation}
	\text{Gower}(i, j) = \frac{ \sum_{k=1}^{N} w_k S_{ijk} }{ \sum_{k=1}^{N} w_k }
\end{equation}
where $N$ is the number of features and $w_k$ is the weight of the $k$-th variable, in this paper the weights are set to one; $S_{ijk}$ is the contribution by the $k$-th variable.
If the $k$-th variable is continuous, $S_{ijk}$ is defined as
\begin{equation}
	S_{ijk} = 1 - \frac{ | x_{ik} - x_{jk} | }{r_k}
\end{equation}
where $x_{ik}$ is the value of $k$-th feature of $i$ and $r_k$ is the range of values for the $k$-th variable.
If the $k$-th variable is categorical, $S_{ijk}$ is 1 if $x_{ik} = x_{jk}$ or 0, otherwise.

\subsection{Model Selection Criterion}
We perform model selection on the validation dataset for two different criterion, namely, \textbf{discrimination} and \textbf{delta}.
Discrimination is the same as the metric used to evaluate the model introduced in Section \ref{sec:evaluation_metrics}. As the model selected by discrimination does not take accuracy into account, we use delta as another selection criterion.
Delta is the difference between accuracy and discrimination. A model selected by delta trades off some fairness for accuracy.



\begin{table*}[h!]
	\centering
	\caption{The prediction results by using different submodels of \textbf{CCB}.} \label{tab:rslt3}
	\begin{adjustbox}{max width=\textwidth}
	\begin{threeparttable}
\begin{tabular}{ccccc}
\toprule
{} &                 adult &                compas &                german &                health \\
{} & acc$|$discri$|$consist & acc$|$discri$|$consist & acc$|$discri$|$consist & acc$|$discri$|$consist \\
\midrule
Model 0        &  $0.7548|0.0019|0.9976$ &  $0.4787|0.0019|0.9926$ &  $0.6950|0.0028|0.9510$ &  $0.8785|0.0000|1.0000$ \\
Model 1        &  $0.7543|0.0000|1.0000$ &  $0.4749|0.0000|1.0000$ &  $0.6850|0.0000|1.0000$ &  $0.8784|0.0002|0.9998$ \\
Reversed Model &  $0.7548|0.0021|0.9977$ &  $0.4768|0.0050|0.9958$ &  $0.6850|0.0310|0.9670$ &  $0.8784|0.0001|0.9999$ \\
Original Model &  $0.7544|0.0002|0.9999$ &  $0.4768|0.0031|0.9960$ &  $0.6950|0.0282|0.9840$ &  $0.8785|0.0002|0.9999$ \\
\bottomrule
\end{tabular}

	\begin{tablenotes}
	\small
	\item acc = accuracy, discri = discrimination, consist = consistency.
	\end{tablenotes}
	\end{threeparttable}
	\end{adjustbox}
\end{table*}

\section{Experimental Results} \label{sec:experiment_results}
\subsection{Results and Analysis}
Table \ref{tab:rslt1} shows the performance metrics on the test set
by using
discrimination as model selection criterion on the validation set. From 
the results, we can see that the proposed method \textbf{CCB} achieves comparable or better performance than baselines . The discrimination scores are pushed close to 0 and consistency scores close to 1. In addition, the proposed method is able to achieve both group fairness (measured by discrimination) and individual fairness (measured by consistency), although the proposed method is targeting at achieving individual fairness. \textbf{LR} that has no fairness constraints achieves almost the best performance in terms of predictive accuracy. However, its predictions have the highest bias. Other methods achieve fairness at the cost of accuracy. \textbf{Rawlsian} as an individual fairness algorithm fails to remove bias from both adult and compas dataset and has limited capability of removing bias for german and health datasets. The reason might be that its implementation using interval chaining is a weak constraint on the model. 
The experimental results also reveal that german and health have far less bias than adult and compas, even \textbf{LR} has very limited discriminative predictions on german and health.

Table \ref{tab:rslt2} shows the results using delta as model selection criterion. As the delta criterion trades some fairness for accuracy, the results shows that the model is not able to achieve the same level of fairness as the ones using discrimination as selection criterion. However, the models are able to achieve higher predictive accuracy. Another observation is that the models are not able to achieve individual fairness by using delta as selection criterion. It is interesting to see that on the two most biased datasets adult and compas, \textbf{CCB} has the most improvement in terms of accuracy compared to other methods by changing the criterion from discrimination to delta. The improvements 
for adult and compas in terms of accuracy are 8.23\% and 31.17\%, respectively.



\begin{table*}[h!]
    \centering
    \caption{Dataset Statistics} \label{tab:data_stats}
    \begin{adjustbox}{max width=0.9\textwidth}
    \begin{threeparttable}

\begin{tabular}{cccccccc} 
\toprule
Major &    \#S &  \#C &      \#G &            \#M &            \#F &          \#AA &          \#NAA \\
\midrule
 Major1 &  6,127 &  16 &  124,716 &  1,927(31.45\%) &  4,200(68.55\%) &  759(12.39\%) &  5,368(87.61\%) \\
 Major2 &   450 &   7 &   23,708 &   338(75.11\%) &   112(24.89\%) &    27(6.00\%) &   423(94.00\%) \\
 Major3 &  2,430 &  11 &   90,819 &  1,942(79.92\%) &   488(20.08\%) &   157(6.46\%) &  2,273(93.54\%) \\
 Major4 &   671 &  10 &   65,396 &   575(85.69\%) &    96(14.31\%) &    66(9.84\%) &   605(90.16\%) \\
 Major5 &  5,110 &  17 &   84,504 &  1,200(23.48\%) &  3,910(76.52\%) &  694(13.58\%) &  4,416(86.42\%) \\
\bottomrule
\end{tabular}

        \begin{tablenotes}
            \scriptsize
            \item \#S total number of students, \#C number of courses for prediction, \#G total number of grades
            \item \#M number of male students, \#F number of female students, \#AA number of African-American students \#NNA number of non-African-American students.
        \end{tablenotes}
    \end{threeparttable}
    \end{adjustbox}
\end{table*}

\begin{table*}[h!]
	\centering
	\caption{Experimental results on educational dataset with gender as sensitive attribute.} \label{tab:ed_rslt1}
	\begin{adjustbox}{max width=0.9\textwidth}}
	\begin{threeparttable}
\begin{tabular}{cccccc}
\toprule
\multirow{2}{2em}{Method} &                  BIOL &                  CEIE &                    CS &                   ECE &                  PSYC \\
 & acc$|$discri$|$consist & acc$|$discri$|$consist & acc$|$discri$|$consist & acc$|$discri$|$consist & acc$|$discri$|$consist \\
\midrule
\textbf{LR}       &  \textbf{0.7662}$|$0.0613$|$0.8152 &  0.6761$|$0.0837$|$0.7451 &  \textbf{0.6628}$|$0.1007$|$0.7569 &  \textbf{0.7545}$|$0.0980$|$0.7655 &  0.7769$|$0.0192$|$0.9578 \\
\textbf{Rawlsian} &  0.5889$|$0.0807$|$0.8120 &  0.6250$|$0.0866$|$0.7052 &  0.5582$|$0.0913$|$0.8301 &  0.6660$|$0.1498$|$0.7036 &  0.7559$|$0.0960$|$0.9396 \\
\textbf{LFR}      &  0.6470$|$0.0369$|$0.9691 &  0.6983$|$0.0518$|$0.9631 &  0.6004$|$0.0228$|$0.9463 &  0.7389$|$0.0273$|$\textbf{0.9912} &  0.7898$|$0.0248$|$0.9865 \\
\textbf{ALFR}     &  0.6802$|$0.0202$|$0.9675 &  \textbf{0.7062}$|$\textbf{0.0240}$|$\textbf{0.9855} &  0.6124$|$\textbf{0.0134}$|$\textbf{0.9821} &  0.7465$|$\textbf{0.0114}$|$0.9783 &  \textbf{0.7903}$|$0.0125$|$0.9878 \\
\textbf{CCB}	  &  0.6663$|$\textbf{0.0089}$|$\textbf{0.9791} &  0.5901$|$0.0334$|$0.9624 &  0.6051$|$0.0257$|$0.9416 &  0.7300$|$0.0279$|$0.9790 &  0.7861$|$\textbf{0.0094}$|$\textbf{0.9954} 
\\
\bottomrule
\end{tabular}
	\begin{tablenotes}
	\small
	\item acc = accuracy, discri = discrimination, consist = consistency.
	\end{tablenotes}
	\end{threeparttable}
	\end{adjustbox}
\end{table*}

\begin{table*}[h!]
	\centering
	\caption{Experimental results on educational dataset with race as sensitive attribute.} \label{tab:ed_rslt2}
	\begin{adjustbox}{max width=0.9\textwidth}
	\begin{threeparttable}
\begin{tabular}{cccccc}
\toprule
\multirow{2}{2em}{Method} &                  BIOL &                  CEIE &                    CS &                   ECE &                  PSYC \\
 & acc$|$discri$|$consist & acc$|$discri$|$consist & acc$|$discri$|$consist & acc$|$discri$|$consist & acc$|$discri$|$consist \\
\midrule
\textbf{LR}       &  \textbf{0.7662}$|$0.1004$|$0.8152 &  0.6761$|$0.1411$|$0.7451 &  \textbf{0.6628}$|$0.1085$|$0.7569 &  \textbf{0.7545}$|$0.1238$|$0.7655 &  0.7769$|$0.0276$|$0.9578 \\
\textbf{Rawlsian} &  0.5854$|$0.1129$|$0.7870 &  0.5849$|$0.3658$|$0.7349 &  0.5561$|$0.1857$|$0.8007 &  0.6999$|$0.1446$|$0.7416 &  0.7608$|$0.0776$|$0.9570 \\
\textbf{LFR}      &  0.6202$|$0.0569$|$0.9051 &  0.7099$|$0.1722$|$\textbf{0.9701} &  0.6107$|$0.0599$|$\textbf{0.9897} &  0.7441$|$0.0800$|$0.9852 &  0.7874$|$0.0172$|$0.9933 \\
\textbf{ALFR}     &  0.6850$|$0.0505$|$0.9504 &  \textbf{0.7274}$|$\textbf{0.0862}$|$0.9688 &  0.6129$|$\textbf{0.0086}$|$0.9715 &  0.7435$|$0.0384$|$\textbf{0.9887} &  \textbf{0.7898}$|$0.0156$|$0.9882 \\
\textbf{CCB}	  &  0.6443$|$\textbf{0.0115}$|$\textbf{0.9767} &  0.6781$|$0.1071$|$0.8798 &  0.5944$|$0.0372$|$0.9824 &  0.7440$|$\textbf{0.0137}$|$0.9842 &  0.7863$|$\textbf{0.0042}$|$\textbf{0.9993} 
\\
\bottomrule
\end{tabular}
	\begin{tablenotes}
	\small
	\item acc = accuracy, discri = discrimination, consist = consistency.
	\end{tablenotes}
	\end{threeparttable}
	\end{adjustbox}
\end{table*}

\subsection{Discrepancy between Submodels}
The \textbf{CCB} model is composed of two gradient contextual bandits. To achieve individual fairness, we put a constraint on the distance of the predictions from the two contextual bandits. If the model is fair, ideally, given an individual the output distributions of the two bandits should be close to each other. If that is the case, we can use this property to do model compression, namely, we just keep one contextual bandit at inference time. In this section, we propose two ways to investigate the discrepancy between the two submodels. 

Table \ref{tab:rslt3} shows the results of using only one submodel for predictions. Model 0 is the model corresponding to sensitive attribute 
0 and Model 1 is the model corresponding to sensitive attribute
1. Reversed Model implies 
that we do predictions for individuals with sensitive attribute 0 using 
Model 1, and for individuals with 
sensitive attribute 1 using Model 0. Original Model implies that we are
predicting for individuals with sensitive attribute 0 using Model 0
and for individuals with sensitive attribute 1 using Model 1. The results show
that there is no significant difference by
using different models for predictions. Thus,  
we can keep any one of the submodels for prediction after it is trained.


\subsection{Convergence Analysis}
\begin{figure} [ht!]
	\centering
	\includegraphics[width=0.95\linewidth]{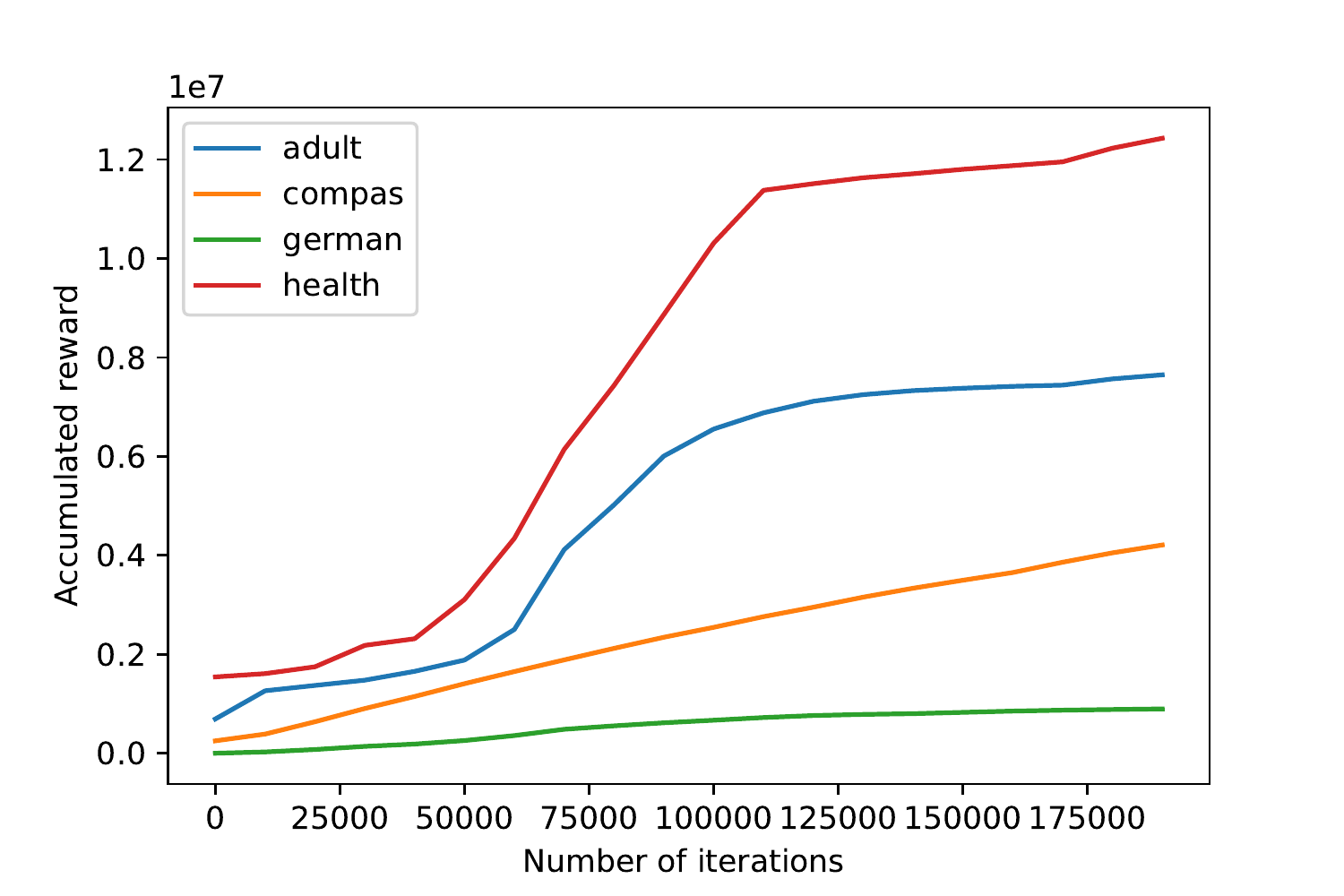}
	\caption{The convergence curves of the proposed \textbf{CCB} model on the four public datasets.} \label{fig:converge}
\end{figure}

Figure \ref{fig:converge} shows the convergence rates of the proposed model on the four public datasets. The X axis is the number of iterations and the Y axis is the accumulated reward. The convergence curves show that the accumulated reward increase monotonically on four datasets. 
The convergence rates are different on different datasets. For the health and adult datasets, the rewards accumulate slowly at the beginning due to exploration and then increase fast; at the end, converge smoothly. On the compas and german datasets, the reward accumulation rates are linear.

\subsection{A Case Study on Educational Setting}
Prior research in the educational domain has focussed on the
development of 
machine learning models for assisting 
students with their learning such as 
performance prediction \cite{elbadrawy2016predicting}, knowledge tracing \cite{piech2015deep} and 
degree pathway recommendations \cite{elbadrawy2016domain}. With the 
increasing adoption of machine learning in education, fairness is a 
big concern. A biased
machine learning model can negatively impact a 
minority group of students. For example, an unfair 
at-risk identification approach 
which predicts some group of 
students to have lower GPAs than others can discourage them. 
To mitigate bias, we apply our proposed model to educational setting and do a case study on student performance prediction.

\subsubsection{Datasets}
The dataset is collected at George Mason University from Fall 2009 to Fall 2019 including the top 5 largest majors, Biology (BIOL), Civil Engineering (CEIE), Computer Science (CS), Electrical Engineering (ECE), and Psychology (PSYC). 
For each course in a major, we build a model to predict if a student is going to fail that course (grade $<$ 3.0). The features fed into the model are students' grades in courses taken prior to the target course. The sensitive attributes include gender (male/female) and race (African-American/Non-African-American). Table \ref{tab:data_stats} shows the statistics of the datasets.

\subsubsection{Experimental Protocol}
Following the  experimental protocol of public dataset, we choose the same baselines and use the same evaluation metrics to evaluate the proposed model on removing bias in the setting of student performance prediction. For model selection, we choose discrimination.

\subsubsection{Experimental Results}
Table \ref{tab:ed_rslt1} shows the results using gender as sensitive attribute. From the results of \textbf{LR}, we can see that different majors have different levels of bias; PSYC has the least biased predictions, while CS has the highest biased predictions. Similar to the results on public datasets, it is interesting to see that \textbf{Rawlsian} is not able to remove bias and in some cases it leads to even more unfair predictions. Table \ref{tab:ed_rslt2} shows the results using race as sensitive attribute. Compared to the results by using gender as sensitive attribute, predictions with race as sensitive attribute is more biased.
For example, in terms of predictions of \textbf{LR}, the discriminations of using gender and race as sensitive attribute are 7.27\% and 10.03\%, respectively.
Overall, the proposed model is able to remove bias for both race and gender. It achieves comparable or even better results than the baselines. Compared to two competitive baselines \textbf{LFR} and \textbf{ALFR} which need to first learn a fair representation and then train a classifier on it, \textbf{CCB} is easier to implement.

\section{Conclusions} \label{sec:conclusion}
In this work, we proposed the concept of metric free individual fairness. 
Traditional definition of individual fairness needs a problem-specific similarity metric, which refrains its adoption. The proposed metric free individual fairness eliminates the requirement of a similarity metric and is easy to implement.
The proposed 
gradient contextual bandit algorithm 
learns a stochastic policy which can be applied to other 
domains such as recommender systems and 
for information retrieval.
We also proposed a  \textbf{CCB} fair algorithm. The experimental results show the effectiveness of the proposed algorithm at removing unfairness.
In the future, we plan to explore applying the proposed gradient contextual bandits to other domains. We also want to explore using different fairness measures as reward signals and see how they influence the performance of the model.

\section{Acknowledgements}
This work was supported by the National Science Foundation grant 1447489 and DUE-1937905. The computational resources was provided by ARGO, a research computing cluster provided by the Office of Research Computing at George Mason University, VA. (URL:http://orc.gmu.edu)

\bibliographystyle{IEEEtran}
\bibliography{refs}

\end{document}